# Mind the Gap: Federated Learning Broadens Domain Generalization in Diagnostic AI Models

Soroosh Tayebi Arasteh (1), Christiane Kuhl (1), Marwin-Jonathan Saehn (1), Peter Isfort (1), Daniel Truhn* (1), Sven Nebelung* (1)

(1) Department of Diagnostic and Interventional Radiology, University Hospital RWTH Aachen, Aachen, Germany.

* D.T. and S.N. are co-senior authors.

## Abstract

Developing robust artificial intelligence (AI) models that generalize well to unseen datasets is challenging and usually requires large and variable datasets, preferably from multiple institutions. In federated learning (FL), a model is trained collaboratively at numerous sites that hold local datasets without exchanging them. So far, the impact of training strategy, i.e., local versus collaborative, on the diagnostic on-domain and off-domain performance of AI models interpreting chest radiographs has not been assessed. Consequently, using 610,000 chest radiographs from five institutions across the globe, we assessed diagnostic performance as a function of training strategy (i.e., local vs. collaborative), network architecture (i.e., convolutional vs. transformer-based), generalization performance (i.e., on-domain vs. off-domain), imaging finding (i.e., cardiomegaly, pleural effusion, pneumonia, atelectasis, consolidation, pneumothorax, and no abnormality), dataset size (i.e., from n=18,000 to 213,921 radiographs), and dataset diversity. Large datasets not only showed minimal performance gains with FL but, in some instances, even exhibited decreases. In contrast, smaller datasets revealed marked improvements. Thus, on-domain performance was mainly driven by training data size. However, off-domain performance leaned more on training diversity. When trained collaboratively across diverse external institutions, AI models consistently surpassed models trained locally for off-domain tasks, emphasizing FL's potential in leveraging data diversity. In conclusion, FL can bolster diagnostic privacy, reproducibility, and off-domain reliability of AI models and, potentially, optimize healthcare outcomes.

**Correspondence:**
Soroosh Tayebi Arasteh, MSc
Department of Diagnostic and Interventional Radiology, University Hospital RWTH Aachen, Pauwelsstr. 30, 52074 Aachen, Germany
Email: sarasteh@ukaachen.de

# Introduction

Artificial Intelligence (AI) is increasingly indispensable for medical imaging[1,2]. Deep learning models can analyze vast amounts of data, extract complex patterns, and assist in the diagnostic workflow[3,4]. In medicine, AI models are applied in various tasks that range from detecting abnormalities[5] to predicting disease progression based on patient data[6]. However, their success hinges on the availability and diversity of available training data. Data drives the learning process, and the performance and generalizability of AI models scale with the amount and variety of data they have been trained on[7,8].

In medical imaging, privacy regulations pose a considerable challenge to data sharing, which limits the ability of researchers and practitioners to access large and diverse datasets crucial for the development of equally robust and performant, i.e., generalizing, AI models. Federated learning (FL) is an approach that may solve this problem[9–14]. FL preserves privacy as the AI models are trained across multiple sites (decentralized), with each site using local datasets that are not exchanged. Critically, sensitive data are stored locally and not transferred, which reduces the risk of data breaches. Practically, the sites collaboratively train the AI model, while local updates, such as gradients, are fed back and aggregated centrally. Once these updates are aggregated, the improved global model is sent back to the sites for subsequent training rounds. While FL is promising in scientific contexts[15], it faces several challenges, including independent and identically distributed (IID) versus non-IID data distributions and variations in image acquisition, processing, and labeling[10]. These challenges may impede the convergence and generalization of the trained AI models[16,17]. AI models trained with IID data (regarding the standardization of labels, image acquisition and image processing routines, cohort characteristics, sample sizes, and imaging feature distributions) perform better, and efforts have been made to harmonize the collaborative training process, benefiting all participating institutions[18–21].

Earlier studies have primarily focused on the impact of IID versus non-IID data settings and on-domain performance in FL strategies[22,23]. Even though the regularly weaker off-domain performance of AI models is increasingly recognized[24–28], there is a substantial gap in our understanding of the impact of FL on the performance of diagnostic AI models. Beyond FL, additional confounding variables are underlying network architectures, dataset size and diversity, and the AI model's outputs, for example, imaging findings.

Our study explores the potential for domain generalization of AI models trained via FL (see **Figure 1**), utilizing over 610,000 chest radiographs from five large datasets. To our knowledge, this is the first analysis of FL applied to the AI-based interpretation of chest radiographs on such a large scale. We conducted all experiments using convolutional and transformer-based network architectures – specifically, the ResNet50[29] and the Vision Transformer (ViT)[30] base models, to assess the potential influence of the underlying architecture[5,31].

We first implement FL across all datasets to study its on-domain effects under non-IID conditions, comparing local versus collaborative training on various datasets. We then assess the off-domain performance of collaboratively trained models, examining the impact of dataset size and diversity. The AI models are collaboratively trained using data from four sites, each with equal contributions, and tested on the fifth site. We also train local models on individual datasets and evaluate their performance on the omitted site. Finally, we test the collaboratively trained models' scalability using each site's full training data sizes. We hypothesize that (i) FL is advantageous in non-IID data conditions and (ii) increased data diversity (secondary to the FL setup) brings about improved off-domain performance.

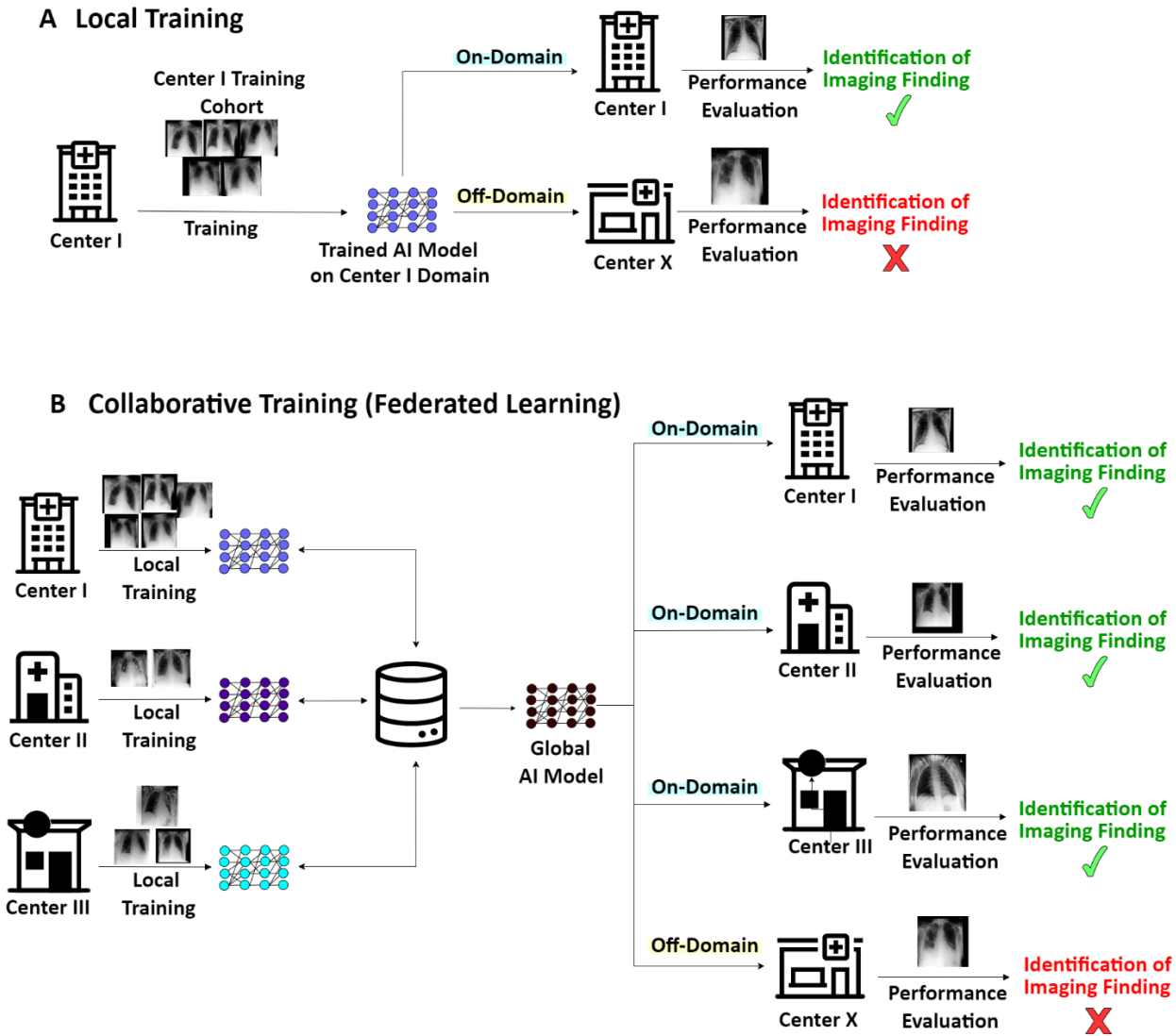

**Figure 1**: **Local and Collaborative Training Processes and the Challenges Associated with Domain Transfer. (A)** *Center I* conventionally trains an AI model to analyze chest radiographs using local data, e.g., bedside chest radiographs of patients in intensive care (supine position, anteroposterior projection). The AI model performs well on test data from the same institution (on-domain), but fails on data from another hospital (*Center X*) that does not operate an intensive care unit but an outpatient clinic with special consultations. Thus, the chest radiographs to be analyzed have been obtained differently (standing position, posteroanterior projection). **(B)** Off-domain performance may be limited following collaborative training, i.e., federated learning.

# Results

## Federated Learning Improves On-Domain Performance in Interpreting Chest Radiographs

On-domain performance varied substantially, often even significantly, between those networks trained locally (at each site) and collaboratively (across the five sites, including the VinDr-CXR[32], ChestX-ray14[33], CheXpert[34], and MIMIC-CXR[35], and PadChest[36] datasets, see **Table 1**) (**Figure 2**). Notably, the VinDr-CXR, ChestX-ray, CheXpert, MIMIC-CXR, and PadChest datasets contained n=15,000, n=86,524, n=128,356, n=170,153, and n=88,480 training radiographs, respectively.

Considering the on-domain performance and all imaging findings, smaller datasets, i.e., VinDr-CXR, ChestX-ray14, and PadChest, were characterized by significantly higher area under the receiver operating characteristic curve (AUROC) values following collaborative training than local training. In contrast, the larger datasets, i.e., CheXpert and MIMIC-CXR, were characterized by similar or slightly lower AUROC values following collaborative training than local training, irrespective of the underlying network architecture (**Figure 2**).

Considering individual imaging findings (or labels), AUROC values varied substantially as a function of dataset, imaging finding, and training strategy (**Tables 2** and **3**). Cardiomegaly, pleural effusion, and no abnormality had consistently (and significantly) higher AUROC values following collaborative training than local training across all datasets. Notably, we found the highest AUROC values for the VinDr-CXR dataset, where collaborative training resulted in close-to-perfect AUROC values for pleural effusion (AUROC=98.6±0.4%) and pneumothorax (AUROC=98.5±0.7%) when using the ResNet50 architecture. Similar observations were made for the ViT architecture. In contrast, for pneumonia, atelectasis, and consolidation, we found similar, or in parts even lower AUROC values following collaborative training (**Tables 2** and **3**), indicating that these imaging findings did not benefit from collaborative training and, consequently, larger datasets.

## Data Diversity is Critical for Enhancing Off-Domain Performance in Federated Learning

We adjusted the training data size to extend our analysis to off-domain performance. We randomly sampled n=15,000 radiographs from the training sets of each dataset for the collaborative training process. We studied five distinct FL scenarios where one dataset was excluded for off-domain assessment and collaborative training was conducted using the remaining four datasets. This approach meant that each FL training process included n=60,000 training radiographs. For comparison, we randomly selected n=60,000 training radiographs from each dataset's training set and used these images to train locally. Subsequently, we evaluated off-domain performance by testing each locally trained network against all other datasets. No overlap existed between the training and test sets in any experiment. We then compared the locally and collaboratively trained models on the same test set. Collaboratively trained models significantly outperformed locally trained models regarding off-domain performance (averaged over all imaging findings) across nearly all datasets (**Tables 4** and **5**).

**Table 1**: **Dataset Characteristics**. Indicated are the included datasets, i.e., VinDr-CXR[28], ChestX-ray14[29], CheXpert[30], MIMIC-CXR[31], and PadChest[32], and their characteristics. Only frontal chest radiographs (both anteroposterior and posteroanterior projections) were used for this study, while lateral projections were disregarded. Multiple radiographs may have been included per patient. N/A: Not available. NLP: Natural Language Processing.

|  | VinDr-CXR | ChestX-ray14 | CheXpert | MIMIC-CXR | PadChest |
|---|---|---|---|---|---|
| Number of Radiographs Total (training set/test set) [n] | 18,000 (15,000 / 3,000) | 112,120 (86,524 / 25,596) | 157,878 (128,356 / 29,320) | 213,921 (170,153 / 43,768 | 110,525 (88,480 / 22,045) |
| Number of Patients (Total) [n] | N/A | 30,805 | 65,240 | 65,379 | 67,213 |
| PATIENT AGE [years] Median Mean ± Standard Deviation Range (minimum, maximum) | 42 54 ± 18 (2, 91) | 49 47 ± 17 (1, 96) | 61 60 ± 18 (18, 91) | N/A N/A N/A | 63 59 ± 20 (1, 105) |
| PATIENT SEX Female / Male [%] Training Set Test Set | 47.8 / 52.2 44.1 / 55.9 | 42.4 / 57.6 41.9 / 58.1 | 41.4 / 58.6 39.0 / 61.0 | N/A N/A | 50.0 / 50.0 48.2 / 51.8 |
| PROJECTIONS [%] anteroposterior posteroanterior | 0.0 100.0 | 40.0 60.0 | 84.5 15.5 | 58.2 41.8 | 17.1 82.9 |
| Country | Vietnam | USA | USA | USA | Spain |
| Contributing Hospitals [n] | 2 | 1 | 1 | 1 | 1 |
| Clinical Setting | N/A | N/A | Inpatient & Outpatient Clinic | Intensive Care Unit | N/A |
| Radiography Systems [n] | ≥ 8 | N/A | N/A | N/A | N/A |
| Labeling Method | Manual | Automatic (NLP) | Automatic (NLP) | Automatic (NLP) | Partially manual, Partially Automatic (NLP) |
| Radiographs with Cardiomegaly [%] | 11.8 | 2.5 | 12.6 | 19.7 | 8.9 |
| Radiographs with Pleural Effusion [%] | 4.1 | 11.9 | 41.3 | 22.6 | 6.3 |
| Radiographs with Pneumonia [%] | 4.0 | 1.3 | 2.5 | 6.5 | 4.7 |
| Radiographs with Atelectasis [%] | 0.8 | 10.3 | 16.7 | 19.9 | 5.6 |
| Radiographs with Consolidation [%] | 1.2 | 4.2 | 6.0 | 4.0 | 1.5 |
| Radiographs with Pneumothorax [%] | 0.4 | 4.7 | 10.3 | 4.6 | 0.4 |
| Radiographs without Abnormality [%] | 70.3 | 53.8 | 10.8 | 37.7 | 32.9 |

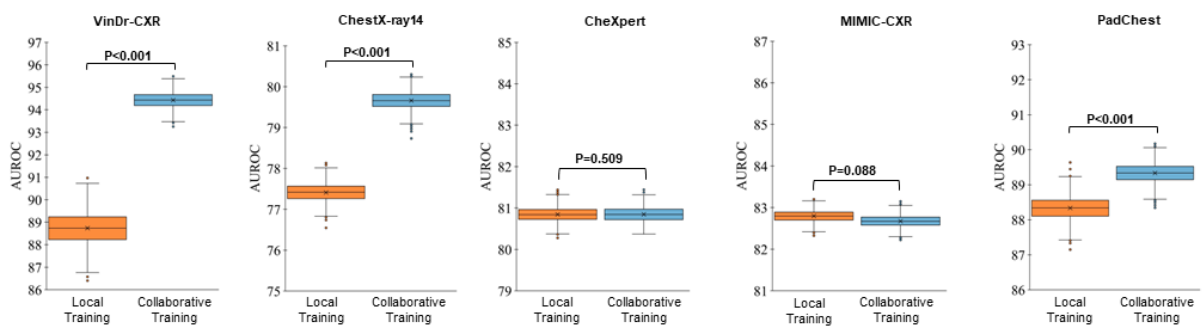

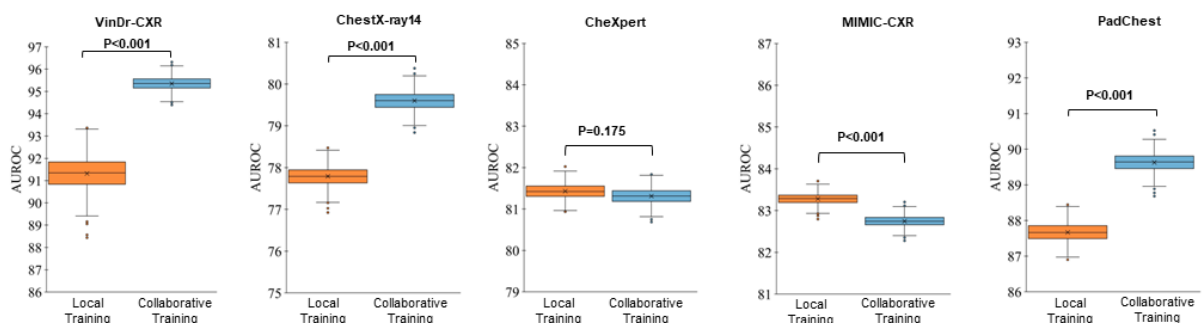

**Figure 2: On-domain Evaluation of Performance – Averaged Over All Imaging Findings.** The results are represented as the area under the receiver operating characteristic curve (AUROC) values averaged over all labeled imaging findings, i.e., cardiomegaly, pleural effusion, pneumonia, atelectasis, consolidation, and pneumothorax, and no abnormalities. "Local Training" (first column, orange) indicates the AUROC values when trained on-domain and locally. "Collaborative Training" (second column, light blue) indicates the corresponding AUROC values when trained on-domain yet collaboratively while including the other datasets (federated learning) as well. The datasets are VinDr-CXR, ChestX-ray14, CheXpert, MIMIC-CXR, and PadChest, with training datasets totaling n=15,000, n=86,524, n=128,356, n=170,153, and n=88,480 chest radiographs, respectively, and test datasets of n=3,000, n=25,596, n=39,824, n=43,768, and n=22,045 chest radiographs, respectively. **(A)** Performance of the ResNet50 architecture, a convolutional neural network. **(B)** Performance of the ViT, a vision transformer. Crosses indicate means, boxes indicate the ranges (first [Q1] to third [Q3] quartile), with the central line representing the median (second quartile [Q2]), whiskers indicate minimum and maximum values, and outliers are indicated with dots. Differences between locally and collaboratively trained models were assessed for statistical significance using bootstrapping, and p-values are indicated.

**Table 2: On-Domain Evaluation of Performance of the Convolutional Neural Network – Individual Imaging Findings.** Performance metrics are indicated as the area under the receiver operating characteristic curve (AUROC) values for each dataset, training strategy (i.e., local or collaborative training), and imaging finding. See **Table 1** for further details on dataset characteristics. Differences between locally and collaboratively trained models were assessed for statistical significance using bootstrapping, and p-values were indicated. Significant differences are indicated in **bold**.

| Dataset | Training Strategy | Cardio-megaly | Pleural Effusion | Pneu-monia | Atelec-tasis | Consoli-dation | Pneumo-thorax | No Abnormality | *Average* |
|---|---|---|---|---|---|---|---|---|---|
| VinDr-CXR | Local | 92.2 ± 0.7 | 93.7 ± 1.4 | 88.3 ± 1.2 | 78.4 ± 3.13 | 88.1 ± 1.9 | 93.3 ± 2.3 | 87.08 ± 0.7 | *88.7 ± 5.2* |
| | Collaborative | 95.3 ± 0.5 | 98.6 ± 0.4 | 89.9 ± 1.0 | 91.2 ± 1.4 | 94.7 ± 1.0 | 98.5 ± 0.7 | 92.9 ± 0.5 | *94.4 ± 3.2* |
| | P-value | **0.001** | **0.001** | 0.896 | **0.001** | **0.001** | **0.003** | **0.001** | **0.001** |
| ChestX-ray14 | Local | 87.5 ± 0.5 | 81.5 ± 0.3 | 68.8 ± 1.1 | 74.7 ± 0.4 | 72.8 ± 0.5 | 84.4 ± 0.4 | 72.2 ± 0.3 | *77.4 ± 6.6* |
| | Collaborative | 89.4 ± 0.5 | 82.6 ± 0.3 | 73.3 ± 1.1 | 77.1 ± 0.4 | 74.7 ± 0.5 | 87.5 ± 0.3 | 73.1 ± 0.3 | *79.7 ± 6.4* |
| | P-value | **0.001** | **0.001** | **0.001** | **0.001** | **0.001** | **0.001** | **0.001** | **0.001** |
| CheXpert | Local | 86.7 ± 0.3 | 87.3 ± 0.2 | 76.4 ± 0.8 | 68.4 ± 0.4 | 74.4 ± 0.5 | 85.5 ± 0.3 | 87.2 ± 0.3 | *80.8 ± 7.1* |
| | Collaborative | 86.7 ± 0.3 | 88.1 ± 0.2 | 73.8 ± 0.9 | 68.8 ± 0.4 | 74.6 ± 0.5 | 86.3 ± 0.3 | 87.7 ± 0.3 | *80.8 ± 7.5* |
| | P-value | 0.443 | **0.001** | **0.001** | 0.864 | 0.681 | **0.001** | **0.001** | 0.509 |
| MIMIC-CXR | Local | 80.9 ± 0.2 | 90.7 ± 0.2 | 73.9 ± 0.5 | 81.7 ± 0.2 | 80.3 ± 0.5 | 86.5 ± 0.4 | 85.4 ± 0.2 | *82.8 ± 5.0* |
| | Collaborative | 78.8 ± 0.2 | 90.9 ± 0.1 | 74.1 ± 0.5 | 81.2 ± 0.2 | 82.2 ± 0.4 | 86.5 ± 0.5 | 85.0 ± 0.2 | *82.7 ± 5.1* |
| | P-value | **0.001** | **0.045** | 0.768 | **0.001** | **0.001** | 0.442 | **0.001** | 0.088 |
| PadChest | Local | 92.2 ± 0.3 | 95.5 ± 0.3 | 84.8 ± 0.7 | 84.4 ± 0.6 | 89.0 ± 0.9 | 86.8 ± 2.0 | 85.8 ± 0.3 | *88.3 ± 3.9* |
| | Collaborative | 92.5 ± 0.2 | 95.9 ± 0.3 | 85.1 ± 0.6 | 84.3 ± 0.6 | 90.0 ± 0.8 | 92.5 ± 1.5 | 85.0 ± 0.3 | *89.3 ± 4.3* |
| | P-value | **0.017** | **0.003** | 0.806 | 0.371 | 0.922 | **0.001** | **0.001** | **0.001** |

## Federated Learning's Off-Domain Performance Scales With Dataset Diversity and Size

To validate whether the collaborative training strategy retains its superior off-domain performance when applied to large and diverse multi-centric datasets, we replicated the off-domain assessment outlined above using the full training size for each dataset following local and collaborative training. We studied five distinct FL scenarios where one dataset was excluded for off-domain assessment, and collaborative training was conducted using the remaining four datasets' full sizes for training (**Figure 3**).

Surprisingly, we observed that all datasets, regardless of their size, were characterized by significantly higher AUROC values following collaborative training than local training (**Figure 3**), irrespective of the underlying network architecture (P<0.001 [ResNet50]; P<0.004 [ViT]). This finding contrasts with our corresponding findings on the on-domain performance (**Figure 2**), which indicates that collaborative training (vs. local training) does not substantially improve performance on larger datasets.

**Table 3: On-Domain Evaluation of Performance of the Vision Transformer – Individual Imaging Findings.** See **Table 2** for further details on table organizaiton.

| Dataset | Training Strategy | Cardio-megaly | Pleural Effusion | Pneu-monia | Atelec-tasis | Consoli-dation | Pneumo-thorax | No Abnormality | *Average* |
|---|---|---|---|---|---|---|---|---|---|
| VinDr-CXR | Local | 95.0 ± 0.5 | 97.2 ± 0.8 | 90.6 ± 0.9 | 86.9 ± 1.7 | 91.1 ± 1.7 | 87.7 ± 3.9 | 90.7 ± 0.6 | *91.3 ± 3.9* |
| | Collaborative | 96.9 ± 0.3 | 98.4 ± 0.5 | 91.2 ± 1.0 | 92.8 ± 1.1 | 96.2 ± 0.7 | 98.1 ± 0.8 | 93.8 ± 0.5 | *95.3 ± 2.6* |
| | P-value | **0.001** | **0.018** | 0.699 | **0.001** | **0.001** | **0.003** | **0.001** | **0.001** |
| ChestX-ray14 | Local | 88.1 ± 0.5 | 81.4 ± 0.3 | 69.5 ± 1.0 | 75.3 ± 0.4 | 73.6 ± 0.5 | 84.3 ± 0.4 | 72.3 ± 0.3 | *77.8 ± 6.4* |
| | Collaborative | 90.2 ± 0.4 | 82.3 ± 0.3 | 73.2 ± 1.0 | 76.9 ± 0.4 | 75.3 ± 0.5 | 86.7 ± 0.3 | 72.5 ± 0.3 | *79.6 ± 6.4* |
| | P-value | **0.001** | **0.001** | **0.001** | **0.001** | **0.001** | **0.001** | 0.919 | **0.001** |
| CheXpert | Local | 87.7 ± 0.3 | 87.6 ± 0.2 | 76.8 ± 0.9 | 68.8 ± 0.4 | 75.1 ± 0.5 | 86.3 ± 0.3 | 87.7 ± 0.3 | *81.4 ± 7.2* |
| | Collaborative | 87.1 ± 0.3 | 88.3 ± 0.2 | 74.9 ± 1.0 | 69.3 ± 0.4 | 75.1 ± 0.5 | 86.6 ± 0.3 | 87.8 ± 0.3 | *81.3 ± 7.3* |
| | P-value | **0.001** | **0.001** | **0.002** | **0.039** | 0.471 | 0.946 | 0.802 | 0.175 |
| MIMIC-CXR | Local | 81.5 ± 0.2 | 90.8 ± 0.1 | 74.4 ± 0.5 | 81.6 ± 0.2 | 82.4 ± 0.4 | 86.8 ± 0.4 | 85.4 ± 0.2 | *83.3 ± 4.8* |
| | Collaborative | 79.2 ± 0.2 | 91.1 ± 0.1 | 73.6 ± 0.5 | 81.5 ± 0.2 | 82.0 ± 0.4 | 87 ± 0.4 | 84.8 ± 0.2 | *82.7 ± 5.2* |
| | P-value | **0.001** | **0.001** | **0.001** | 0.264 | 0.077 | 0.686 | **0.001** | **0.001** |
| PadChest | Local | 91.9 ± 0.3 | 95.3 ± 0.3 | 83.0 ± 0.7 | 81.6 ± 0.6 | 88.1 ± 0.8 | 89.5 ± 1.3 | 84.3 ± 0.3 | *87.7 ± 4.7* |
| | Collaborative | 92.8 ± 0.2 | 96.0 ± 0.3 | 84.5 ± 0.6 | 85.1 ± 0.6 | 91.0 ± 0.6 | 92.6 ± 1.3 | 85.4 ± 0.3 | *89.6 ± 4.3* |
| | P-value | **0.001** | **0.001** | **0.002** | **0.001** | **0.001** | **0.009** | **0.001** | **0.001** |

**Table 4: Off-domain Evaluation of Performance of the Convolutional Neural Network – Standardized Training Data Sizes.** Following local or collaborative training and testing on another dataset, performance was evaluated by averaging AUROC values over all imaging findings. Collaborative training used the remaining four datasets, each contributing n=15,000 training radiographs. Notably, the VinDr-CXR local model was trained using all available images (*), i.e., n=15,000, while the local models of the other datasets were trained using n=60,000 training radiographs. Differences between locally and collaboratively trained models were assessed for statistical significance using bootstrapping, and p-values were indicated. Significant differences are indicated in **bold**. Data are presented as AUROC value (p-value). Abbreviation: OND = On-Domain.

| Training Strategy | Train on: Dataset [Size] | Test on: VinDr-CXR | ChestX-ray14 | CheXpert | MIMIC-CXR | PadChest |
|---|---|---|---|---|---|---|
| Local Training | VinDr-CXR [n=15000] (*) | OND | 64.2 ± 5.0 **(0.001)** | 67.5 ± 10.4 **(0.001)** | 71.2 ± 6.2 **(0.001)** | 75.8 ± 8.1 **(0.001)** |
| | ChestX-ray14 [n=60000] | 84.6 ± 6.6 (**0.005**) | OND | 73.6 ± 7.8 **(0.001)** | 74.6 ± 7.4 **(0.001)** | 80.4 ± 7.6 **(0.001)** |
| | CheXpert [n=60000] | 85.6 ± 6.9 (**0.020**) | 74.0 ± 5.6 (0.339) | OND | 76.9 ± 7.1 (**0.006**) | 81.2 ± 8.0 **(0.001)** |
| | MIMIC-CXR [n=60000] | 86.9 ± 6.3 (0.553) | 73.4 ± 4.2 (**0.008**) | 76.5 ± 7.3 **(0.001)** | OND | 82.4 ± 6.3 (0.794) |
| | PadChest [n=60000] | 84.7 ± 6.6 (**0.012**) | 70.7 ± 6.9 **(0.001)** | 73.0 ± 8.5 **(0.001)** | 74.5 ± 7.3 **(0.001)** | OND |
| Collaborative Training | All Datasets [n=4 x 15000] | 87.0 ± 6.0 | 73.9 ± 5.0 | 74.5 ± 8.6 | 76.6 ± 6.2 | 82.8 ± 6.7 |

**Table 5: Off-domain Evaluation of Performance of the Vision Transformer – Standardized Training Data Sizes.** Table organization as in **Table 4** above.

| Training Strategy | Train on: Dataset [Size] | Test on: VinDr-CXR | ChestX-ray14 | CheXpert | MIMIC-CXR | PadChest |
|---|---|---|---|---|---|---|
| Local Training | VinDr-CXR [n=15000] (*) | OND | 66.4 ± 5.9 **(0.001)** | 69.3 ± 10.1 **(0.001)** | 73.4 ± 6.3 **(0.001)** | 79.6 ± 6.6 **(0.001)** |
| | ChestX-ray14 [n=60000] | 85.9 ± 6.7 (**0.001**) | OND | 75.0 ± 7.6 **(0.001)** | 76.5 ± 6.1 **(0.001)** | 82.9 ± 6.6 **(0.001)** |
| | CheXpert [n=60000] | 85.3 ± 8.3 (**0.001**) | 75.3 ± 7.6 (**0.039**) | OND | 78.0 ± 6.6 **(0.001)** | 82.1 ± 7.9 **(0.001)** |
| | MIMIC-CXR [n=60000] | 90.0 ± 5.4 (**0.008**) | 75.0 ± 4.6 (0.468) | 77.6 ± 7.1 **(0.001)** | OND | 85.1 ± 5.4 (0.747) |
| | PadChest [n=60000] | 88.8 ± 5.0 **(0.001)** | 72.6 ± 5.6 **(0.001)** | 74.3 ± 7.9 **(0.001)** | 76.7 ± 6.2 **(0.001)** | OND |
| Collaborative Training | All Datasets [n=4 x 15000] | 91.1 ± 4.2 | 75.0 ± 6.0 | 76.5 ± 7.8 | 78.7 ± 5.8 | 85.2 ± 5.7 |

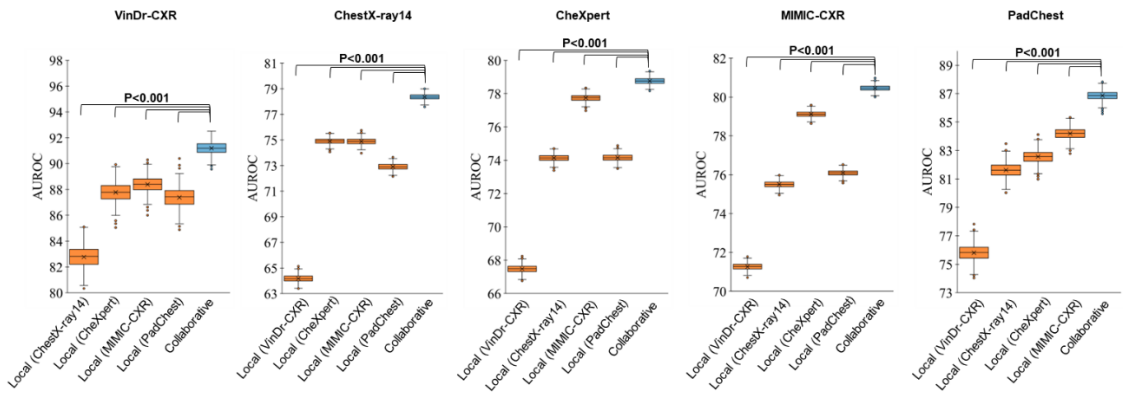

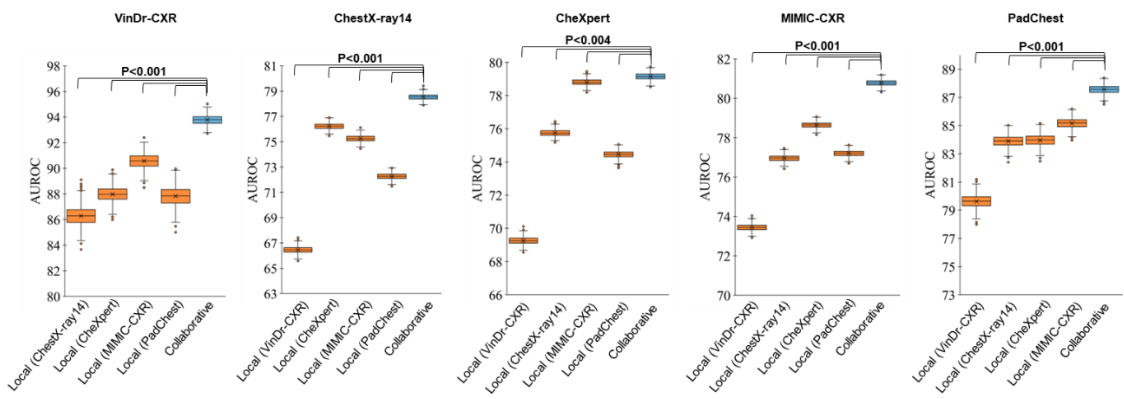

**Figure 3**: **Off-domain Evaluation of Performance – Averaged Over All Imaging Findings.** The results are represented as AUROC values averaged over all labeled imaging findings. The dataset outlined above each subpanel provides the test set, while the first four columns (orange) indicate the AUROC values when trained locally on other datasets, i.e., off-domain. The fifth column (light blue) indicates the corresponding AUROC values when trained off-domain yet collaboratively while including all four datasets (federated learning). Otherwise, the figure is organized as **Figure 2**. Mind the different y-axis scales.

# Discussion

In this study, we examined the impact of federated learning on domain generalization for an AI model that interprets chest radiographs. Utilizing over 610,000 chest radiographs from five datasets from the US, Europe, and Asia, we analyzed which factors influence the off-domain performance of locally versus collaboratively trained models. Beyond training strategies, dataset characteristics, and imaging findings, we also studied the impact of the underlying network architecture, i.e., a convolutional neural network (ResNet50[29]) and a vision transformer (12-layer ViT[30]).

We examined the on-domain performance, i.e., the AI model's performance on data from those institutions that provided data for the initial training, as a function of training strategy using the full training datasets of all five institutions. The collaborative training process unfolded within a predominantly non-IID data setting, with each institution providing inherently variable training images regarding the clinical situation, labeling method, and patient demographics. Previous studies have indicated that FL using non-IID data settings may yield suboptimal results for AI models[14,18,19,19,20,37]. Our results complement these earlier findings as we observed that the degree to which non-IID settings affect the AI models' performance depends on the training data quantity. Institutions with access to large training datasets, such as MIMIC-CXR[35] and CheXpert[34], containing n=170,153 and n=128,356 training radiographs, respectively, demonstrated the least performance gains secondary to FL. In contrast, the VinDr-CXR[32] dataset, with only n=15,000 training radiographs, had the largest performance gains. Our findings confirm that training data size is the primary determinant of on-domain model performance following collaborative training in non-IID data settings, representing most clinical situations.

Consequently, we examined FL and its effects on off-domain performance, i.e., the AI models' performance on unseen data from institutions that did not partake in the initial training[25–27,38]. First, to study if factors other than data size would impact off-domain performance, we compared the off-domain performance of the AI model trained locally when each dataset's size matched the combined dataset size used for collaborative training. We found significant improvements in AUROC values in most collaborative and local training strategies. This finding suggests that -contrary to on-domain performance, which is affected by dataset size- off-domain performance is influenced by the diversity of the training data. Notably, the MIMIC-CXR[35] and the CheXpert[34] datasets used the same labeling approach, which may explain why the AI models trained on either of these datasets performed at least as well as their counterparts trained collaboratively. Second, we evaluated the off-domain performance using the complete training datasets to determine the scalability of FL. The collaboratively trained AI models consistently outperformed their locally trained counterparts regarding average AUROC values across all imaging findings. Thus, FL enhances the off-domain performance by leveraging dataset diversity and size.

To study the effect of the underlying network architecture, we assessed convolutional and transformer-based networks, namely ResNet50 and ViT base models. Despite marginal differences, both architectures displayed comparable performance in interpreting chest radiographs[39].

Surprisingly, the diagnostic performance regarding pneumonia, atelectasis, and consolidation did not benefit from larger datasets (following collaborative training) as opposed to cardiomegaly, pleural effusion, and no abnormality. This finding is surprising in light of the variable, yet still relatively low prevalence of pneumonia (1.3% – 6.5%), atelectasis (0.8% - 19.9%), and consolidation (1.2% - 6.0%) across the datasets. Intuitively, one would expect the diagnostic performance to benefit from more and more variable datasets. While the substantial variability in image and label quality may be responsible, further studies are necessary to corroborate or refute this finding.

Our study has limitations: First, we recognize that our collaborative training was conducted within a single institution's network. By segregating the computing entity for each (virtual) site participating in the AI model's collaborative training, we emulated a practical scenario where network parameters from various sites converge at a central server for aggregation. In real-world scenarios, collaborative training across institutions translates to disparate physical locations where network latency and computational resources affect procedural efficiency. Importantly, diagnostic performance metrics will not be affected by these factors. Second, we had to rely on the label quality and consistency provided along with the radiographs by the dataset providers, which may be problematic[40]. Third, although our study used numerous real-world datasets, it exclusively focused on chest radiographs. In the future, AI models that assess other imaging and non-imaging features as surrogates of health outcomes should be studied.

In conclusion, our multi-institutional study of the AI-based interpretation of chest radiographs using variable dataset characteristics pinpoints the potential of federated learning in (i) facilitating privacy-preserving cross-institutional collaborations, (ii) leveraging the potential of publicly available data resources, and (iii) enhancing the off-domain reliability and efficacy of diagnostic AI models. Besides promoting transparency and reproducibility, the broader future implementation of sophisticated collaborative training strategies may improve off-domain deployability and performance and, thus, optimize healthcare outcomes.

# Materials and Methods

## Ethics Statement

The study was performed in accordance with relevant local and national guidelines and regulations and approved by the Ethical Committee of the Faculty of Medicine of RWTH Aachen University (Reference No. EK 028/19).

## Patient Cohorts

Our study includes 612,444 frontal chest radiographs from various institutions, i.e., the VinDr-CXR[32], ChestX-ray14[33], CheXpert[34], MIMIC-CXR[35], and the PadChest[36] datasets. The median patient age was 58, with a mean (± standard deviation) of 56 (± 19) years. Patient ages ranged from 1 to 105 years. Beyond dataset demographics, we provide additional dataset characteristics, such as labeling systems, label distributions, gender, and imaging findings, in **Table 1**.

The VinDr-CXR[32] dataset, collected from 2018 to 2020, was provided by two large hospitals in Vietnam and includes 18,000 frontal chest radiographs, all manually annotated by radiologists on a binary classification scheme to indicate an imaging finding's presence or absence. For the training set, each chest radiograph was independently labeled by three radiologists, while the test set labels represent the consensus among five radiologists[32]. The official training and test sets comprise n=15,000 and n=3,000 images, respectively.

The ChestX-ray14[33] dataset, gathered from the National Institutes of Health Clinical Center (US) between 1992 and 2015, includes 112,120 frontal chest radiographs from 30,805 patients. Labels were automatically generated based on the original radiologic reports using natural language processing (NLP) and rule-based labeling techniques with keyword matching. Imaging findings were also indicated on a binary basis. The official training and test sets contain n=86,524 and n=25,596 radiographs, respectively.

The CheXpert[34] dataset from Stanford Hospital (US) features n=157,878 frontal chest radiographs from 65,240 patients. Obtained from inpatient and outpatient care patients between 2002 and 2017, the radiographs were automatically labeled based on the original radiologic reports using an NLP-based labeler with keyword matching. The labels contained four classes, namely "positive", "negative", "uncertain", and "not mentioned in the reports", with the "uncertain" label capturing both diagnostic uncertainty and report ambiguity[34]. This dataset does not offer official training or test set divisions.

The MIMIC-CXR[35] dataset includes n=210,652 frontal chest radiographs from 65,379 patients in intensive care at the Beth Israel Deaconess Medical Center Emergency Department (US) between 2011 and 2016. The radiographs were automatically labeled based on the original radiologic reports utilizing the NLP-based labeler of the CheXpert[34] dataset detailed above. The official test set consists of n=2,844 frontal images.

The PadChest[36] dataset contains n=110,525 frontal chest radiographs from 67,213 patients. These images were obtained at the San Juan Hospital (Spain) from 2009 to 2017. 27% of the radiographs were manually annotated using a binary classification by trained radiologists, while the remaining 73% were labeled automatically using a supervised NLP method to determine the presence or absence of an imaging finding [36].

## Experimental Design

To maintain benchmarking consistency, we standardized the test sets across all experiments. Specifically, we retained the original test sets of the VinDr-CXR and ChestX-ray14 datasets, consisting of n=3,000 and n=25,596 radiographs, respectively. For the other datasets, we randomly selected a held-out subset comprising 20% of the radiographs, i.e., n=29,320 (CheXpert), n=43,768 (MIMIC-CXR), and n=22,045 (PadChest), respectively. Importantly, there was no patient overlap between the training and test sets.

We assessed the AI models' on-domain and off-domain performance in interpreting chest radiographs. On-domain performance refers to applying the AI model on a held-out test set from an institution that participated in the initial training phase through single-institutional local training or multi-institutional collaborative training (i.e., federated learning). Conversely, off-domain performance involves applying the AI model to a test set from an institution that did not participate in the initial training phase, regardless of whether the training was local or collaborative.

*Pre-Processing*

The diagnostic labels of interest included cardiomegaly, pleural effusion, pneumonia, atelectasis, consolidation, pneumothorax, and no abnormality. To align with previous studies[13,25,41,42], we implemented a binary multi-label classification system, enabling each radiograph to be assigned a positive or negative class for each imaging finding. As a result, labels from datasets with non-binary labeling systems were converted to a binary classification system. Specifically, for datasets with certainty levels in their labels, i.e., CheXpert and MIMIC-CXR, classes labeled as "certain negative" and "uncertain" were summarized as "negative", while only the "certain positive" class was treated as "positive". To ensure consistency across datasets, we implemented a standardized multi-step image pre-processing strategy: First, the radiographs were resized to the dimension of $224 \times 224$ pixels. Second, min-max feature scaling, as proposed by Johnson et al.[35], was implemented. Third, to improve image contrast, histogram equalization was applied[13,35].

*Federated Learning*

When designing our FL study setup, we followed the FedAvg algorithm proposition by McMahan et al.[11]. Consequently, each of the five institutions was tasked with carrying out a local training session, after which the network parameters, i.e., the weights and biases, were sent to a secure server. This server then amalgamated all local parameters, resulting in a unified set of global parameters. For our study, we set one round to be equivalent to a single training epoch utilizing the full local dataset. Subsequently, each institution received a copy of the global network from the server for another iteration of local training. This iterative process was sustained until a point of convergence was reached for the global network. Critically, each institution had no access to the other institutions' training data or network parameters. They only received an aggregate network without any information on the contributions of other participating institutions to the global network. Following the convergence of the training phase for the global classification network, each institution had the opportunity to retain a copy of the global network for local utilization on their respective test data[12,14].

# DL Network Architecture and Training

Convolutional Neural Network: We utilized a 50-layer implementation of the ResNet architecture (ResNet50), as introduced by He et al.[29], for our convolutional-based network architecture. The initial layer consisted of a $(7 \times 7)$ convolution, generating an output image with 64 channels. The network inputs were $(224 \times 224 \times 3)$ images, processed in batches of 128. The final linear layer was designed to reduce the $(2048 \times 1)$ output feature vectors to the requisite number of imaging findings for each comparison. A binary sigmoid function converted output predictions into individual class probabilities. The optimization of ResNet50 models was performed using the Adam[43] optimizer with learning rates set at $1 \times 10^{-4}$. The network comprised approximately 23 million trainable parameters.

Transformer Network: We adopted the original 12-layer vision transformer (ViT) implementation, as proposed by Dosovitskiy et al.[30], as our transformer-based network architecture. The network was fed with $(224 \times 224 \times 3)$ images in batches of size 32. The embedding layer consisted of a $(16 \times 16)$ convolution with a stride of $(16 \times 16)$, followed by a positional embedding layer, which yielded an output sequence of vectors with a hidden layer size of 768. These vectors were supplied to a standard transformer encoder. A Multi-Layer Perceptron with a size of 3,072 served as the classification head. As with the ResNet50, a binary sigmoid function was used to transform the output predictions into individual class probabilities. The ViT models were optimized using the AdamW[44] optimizer with learning rates set at $1 \times 10^{-5}$. The network comprised approximately 86 million trainable parameters.

All models were pre-trained on the ImageNet-21K[45] dataset with approximately 21,000 categories. Data augmentation strategies included applying random rotation within the range of $[-10, 10]$ degrees and flipping[11]. We used binary weighted Cross-Entropy with inverted class frequencies of the training data as our loss function.

## Evaluation Metrics and Statistical Analysis

We analyzed the AI models using Python (v3) and the SciPy and NumPy packages. The primary evaluation metric was the area under the receiver operating characteristic curve (AUROC), supplemented by additional evaluation metrics such as accuracy, specificity, and sensitivity (**Supplementary Tables S1–S3**). The thresholds were chosen according to Youden's criterion[46]. We employed bootstrapping[47] with repetitions and 1,000 redraws in the test sets to determine the statistical spread and whether AUROC values differed significantly. Multiplicity-adjusted p-values were determined based on the false discovery rate to account for multiple comparisons, and the family-wise alpha threshold was set to 0.05.

## Code Availability

All source codes for training and evaluation of the deep neural networks, data augmentation, image analysis, and pre-processing are publicly available at https://github.com/tayebiarasteh/FLdomain. All code for the experiments was developed in Python v3.9 using the PyTorch v2.0 framework.

## Data Availability

The accessibility of the utilized data in this study is as follows: The ChestX-ray14 and PadChest datasets are publicly available via https://www.v7labs.com/open-datasets/chestx-ray14 and https://bimcv.cipf.es/bimcv-projects/padchest/, respectively. The VinDr-CXR and MIMIC-CXR datasets are restricted-access resources, which can be accessed from PhysioNet by agreeing to the respective data protection requirements under https://physionet.org/content/vindr-cxr/1.0.0/ and https://physionet.org/content/mimic-cxr-jpg/2.0.0/, respectively. The CheXpert dataset may be requested at https://stanfordmlgroup.github.io/competitions/chexpert/.

## Hardware

The hardware used in our experiments were Intel CPUs with 18 cores and 32 GB RAM and Nvidia RTX 6000 GPU with 24 GB memory.

## Funding Sources

STA is funded and supported by the Radiological Cooperative Network (RACOON) under the German Federal Ministry of Education and Research (BMBF) grant number 01KX2021. SN and DT were supported by grants from the Deutsche Forschungsgemeinschaft (DFG) (NE 2136/3-1, TR 1700/7-1). DT is supported by the German Federal Ministry of Education (TRANSFORM LIVER, 031L0312A; SWAG, 01KD2215B) and the European Union's Horizon Europe and innovation programme (ODELIA [Open Consortium for Decentralized Medical Artificial Intelligence], 101057091).

## Author Contributions

STA, DT, and SN designed the study and performed the formal analysis. The manuscript was written by STA and reviewed and corrected by DT and SN. The experiments were performed by STA. The software was developed by STA. The statistical analyses were performed by STA, DT, and SN. CK, MJS, PI, DT, and SN provided clinical expertise. STA and DT provided technical expertise. STA pre-processed the data. All authors read the manuscript and agreed to the submission of this paper.

## Competing Interests

DT holds shares in StratifAI GmbH. The other authors declare no competing interests.

# Supplementary Information

**Table S1: On-Domain Evaluation.** The results are represented as the accuracy, sensitivity, and specificity percentage values averaged over all imaging findings, i.e., cardiomegaly, pleural effusion, pneumonia, atelectasis, consolidation, and pneumothorax as well as no abnormality for each dataset, utilizing the ResNet50 architecture (as the prototypical Convolutional Neural Network) and the ViT architecture (as the prototypical Transformer Network). Two training strategies were used, i.e., local training and collaborative training (i.e., federated learning). The datasets employed in this study were the VinDr-CXR, ChestX-ray14, CheXpert, MIMIC-CXR, and PadChest datasets with n=15,000, n=86,524, n=128,356, n=170,153, and n=88,480 training radiographs, and n=3,000, n=25,596, n=39,824, n=43,768, and n=22,045 test radiographs.

| Test Dataset | Training Strategy | Convolutional Neural Network | | | Transformer Network | | |
|---|---|---|---|---|---|---|---|
| | | Accuracy | Sensitivity | Specificity | Accuracy | Sensitivity | Specificity |
| VinDr-CXR | Local | 84.9 ± 3.9 | 80.9 ± 8.7 | 83.8 ± 6.1 | 84.7 ± 7.2 | 84.8 ± 8.9 | 84.4 ± 7.5 |
| | Collaborative | 88.4 ± 5.2 | 88.3 ± 5.9 | 87.4 ± 6.1 | 89.1 ± 4.9 | 89.0 ± 5.2 | 88.8 ± 5.3 |
| ChestX-ray14 | Local | 70.7 ± 6.4 | 70.7 ± 9.8 | 71.8 ± 7.5 | 70.2 ± 7.5 | 72.0 ± 9.3 | 71.0 ± 8.8 |
| | Collaborative | 71.6 ± 7.8 | 74.1 ± 8.7 | 72.4 ± 8.8 | 72.3 ± 6.3 | 73.1 ± 9.6 | 73.4 ± 7.4 |
| CheXpert | Local | 71.2 ± 9.4 | 78.3 ± 5.7 | 69.9 ± 10.3 | 73.4 ± 8.2 | 76.7 ± 7.7 | 72.5 ± 8.5 |
| | Collaborative | 73.5 ± 10.5 | 75.7 ± 9.6 | 72.5 ± 11.9 | 74.6 ± 7.7 | 75.4 ± 8.0 | 74.1 ± 8.0 |
| MIMIC-CXR | Local | 74.0 ± 5.6 | 77.9 ± 7.0 | 73.2 ± 6.3 | 74.0 ± 5.7 | 78.9 ± 6.7 | 72.9 ± 6.5 |
| | Collaborative | 74.4 ± 5.6 | 77.7 ± 6.4 | 73.6 ± 5.9 | 73.5 ± 6.6 | 78.3 ± 6.3 | 72.8 ± 7.3 |
| PadChest | Local | 80.8 ± 4.4 | 83.0 ± 6.0 | 80.3 ± 5.0 | 78.5 ± 5.7 | 83.2 ± 6.7 | 78.0 ± 6.1 |
| | Collaborative | 80.8 ± 5.5 | 84.4 ± 5.3 | 80.3 ± 6.1 | 81.6 ± 5.2 | 84.5 ± 6.0 | 81.0 ± 6.0 |

**Table S2: Off-domain Evaluation of Performance of the Convolutional Neural Network – Standardized Training Data Sizes.** Data are accuracy, sensitivity, and specificity, averaged over all imaging findings when trained locally or collaboratively (i.e., utilizing federated learning) and tested on another dataset. The collaborative training strategy used the remaining four datasets, each contributing n=15,000 training radiographs. Notably, the VinDr-CXR local model was trained using all available radiographs (*), i.e., n=15,000, while the local models of the other datasets were trained using n=60,000 radiographs. The test sets included n=3,000 (VinDr-CXR dataset), n=25,596 (ChestX-ray14 dataset), n=39,824 (CheXpert dataset), n=43,768 (MIMIC-CXR dataset), and n=22,045 (PadChest dataset) radiographs, respectively. OND: On-Domain.

| Train on: | | Evaluation Metric | Test on: | | | | |
|---|---|---|---|---|---|---|---|
| Training Strategy | Dataset [Size] | | VinDr-CXR | ChestX-ray14 | CheXpert | MIMIC-CXR | PadChest |
| Local Training | VinDr-CXR [n=15000] (*) | Accuracy | OND | 54.3 ± 10.4 | 64.0 ± 12.6 | 63.0 ± 6.7 | 71.3 ± 7.3 |
| | | Sensitivity | | 68.9 ± 11.4 | 65.4 ± 16.9 | 71.4 ± 9.3 | 71.0 ± 12.2 |
| | | Specificity | | 53.6 ± 14.6 | 62.5 ± 14.8 | 61.5 ± 9.0 | 70.6 ± 7.7 |
| | ChestX-ray14 [n=60000] | Accuracy | 79.2 ± 7.7 | OND | 65.0 ± 10.8 | 67.1 ± 7.1 | 75.2 ± 7.6 |
| | | Sensitivity | 76.8 ± 8.0 | | 72.9 ± 7.3 | 71.9 ± 9.1 | 74.1 ± 10.3 |
| | | Specificity | 79.4 ± 8.0 | | 63.7 ± 12.2 | 66.0 ± 7.6 | 74.5 ± 8.3 |
| | CheXpert [n=60000] | Accuracy | 79.1 ± 9.8 | 66.6 ± 7.1 | OND | 71.1 ± 6.1 | 76.4 ± 7.2 |
| | | Sensitivity | 78.5 ± 9.8 | 69.8 ± 9.2 | | 71.4 ± 11.4 | 74.1 ± 10.6 |
| | | Specificity | 78.6 ± 10.3 | 67.3 ± 9.2 | | 70.7 ± 6.9 | 76.2 ± 7.4 |
| | MIMIC-CXR [n=60000] | Accuracy | 81.3 ± 6.7 | 67.6 ± 8.2 | 68.3 ± 9.8 | OND | 74.4 ± 8.4 |
| | | Sensitivity | 78.8 ± 9.2 | 67.8 ± 9.7 | 74.0 ± 6.5 | | 78.3 ± 8.7 |
| | | Specificity | 80.5 ± 8.1 | 68.3 ± 9.8 | 67.4 ± 10.8 | | 73.5 ± 9.1 |
| | PadChest [n=60000] | Accuracy | 77.9 ± 9.9 | 62.9 ± 10.6 | 68.5 ± 9.7 | 66.8 ± 7.2 | OND |
| | | Sensitivity | 77.9 ± 8.9 | 68.8 ± 9.8 | 68.2 ± 12.7 | 72.1 ± 9.3 | |
| | | Specificity | 77.3 ± 10.3 | 63.2 ± 12.6 | 67.5 ± 10.6 | 65.6 ± 8.2 | |
| Collaborative Training | All Datasets [n=4 x 15000] | Accuracy | 82.5 ± 6.5 | 65.9 ± 9.0 | 69.1 ± 10.0 | 67.1 ± 6.8 | 73.3 ± 9.1 |
| | | Sensitivity | 77.0 ± 9.3 | 70.3 ± 8.2 | 70.2 ± 11.5 | 75.2 ± 6.4 | 79.2 ± 9.4 |
| | | Specificity | 82.5 ± 6.8 | 66.4 ± 10.7 | 68.0 ± 11.2 | 65.8 ± 7.3 | 72.8 ± 9.1 |

**Table S3: Off-domain Evaluation of Performance of the Vision Transformer – Standardized Training Data Sizes.** Data organization as in **Table S2**.

| Train on: | | Evaluation Metric | Test on: | | | | |
|---|---|---|---|---|---|---|---|
| Training Strategy | Dataset [Size] | | VinDr-CXR | ChestX-ray14 | CheXpert | MIMIC-CXR | PadChest |
| Local Training | VinDr-CXR [n=15000] (*) | Accuracy | OND | 54.0 ± 11.7 | 64.1 ± 13.8 | 63.4 ± 7.2 | 69.5 ± 9.7 |
| | | Sensitivity | | 72.8 ± 12.2 | 67.9 ± 17.4 | 75.0 ± 7.6 | 79.2 ± 8.0 |
| | | Specificity | | 53.3 ± 16.3 | 62.2 ± 16.6 | 61.4 ± 8.7 | 68.7 ± 9.7 |
| | ChestX-ray14 [n=60000] | Accuracy | 79.4 ± 10.9 | OND | 67.7 ± 8.9 | 67.1 ± 6.5 | 74.7 ± 8.3 |
| | | Sensitivity | 78.2 ± 9.1 | | 71.7 ± 8.1 | 74.6 ± 7.5 | 78.0 ± 8.6 |
| | | Specificity | 79.8 ± 11.4 | | 66.8 ± 9.8 | 65.9 ± 7.2 | 74.1 ± 8.9 |
| | CheXpert [n=60000] | Accuracy | 82.4 ± 6.2 | 67.77 ± 8.6 | OND | 71.3 ± 6.1 | 75.8 ± 6.5 |
| | | Sensitivity | 76.7 ± 13.7 | 71.3 ± 10.6 | | 73.2 ± 10.2 | 76.8 ± 11.5 |
| | | Specificity | 82.6 ± 6.8 | 68.6 ± 10.5 | | 70.5 ± 7.2 | 75.3 ± 7.0 |
| | MIMIC-CXR [n=60000] | Accuracy | 84.1 ± 5.9 | 67.8 ± 6.4 | 69.6 ± 8.7 | OND | 77.1 ± 6.4 |
| | | Sensitivity | 82.1 ± 9.6 | 69.9 ± 7.6 | 74.5 ± 6.3 | | 80.1 ± 7.9 |
| | | Specificity | 83.1 ± 7.9 | 68.7 ± 7.9 | 68.6 ± 9.6 | | 76.5 ± 6.8 |
| | PadChest [n=60000] | Accuracy | 82.3 ± 6.3 | 64.0 ± 9.1 | 67.3 ± 10.6 | 67.2 ± 7.9 | OND |
| | | Sensitivity | 82.1 ± 7.0 | 70.3 ± 8.8 | 71.2 ± 12.1 | 75.1 ± 9.8 | |
| | | Specificity | 82.1 ± 6.5 | 64.6 ± 11.2 | 65.9 ± 11.7 | 65.8 ± 9.2 | |
| Collaborative Training | All Datasets [n=4 x 15000] | Accuracy | 83.5 ± 6.1 | 66.1 ± 8.4 | 69.3 ± 10.3 | 69.7 ± 6.1 | 76.5 ± 6.1 |
| | | Sensitivity | 84.7 ± 7.7 | 71.5 ± 9.5 | 72.5 ± 10.0 | 75.0 ± 6.4 | 80.9 ± 6.8 |
| | | Specificity | 83.2 ± 6.5 | 67.2 ± 10.8 | 68.3 ± 11.4 | 68.9 ± 6.8 | 75.9 ± 6.4 |